# Predicting Concentration Levels of Air Pollutants by Transfer Learning and Recurrent Neural Network

Iat Hang Fong[a], Tengyue Li[a], Simon Fong[a], Raymond K. Wong[b], Antonio J. Tallón-Ballesteros[c,*]

[a]Department of Computer and Information Science, University of Macau, Taipa, China.
[b]Faculty of Engineering, University of New South Wales, Sydney. Australia.
[c]Department of Languages and Computer Systems, University of Seville, Seville, Spain.

**Abstract**

Air pollution (AP) poses a great threat to human health, and people are paying more attention than ever to its prediction. Accurate prediction of AP helps people to plan for their outdoor activities and aids protecting human health. In this paper, long-short term memory (LSTM) recurrent neural networks (RNNs) have been used to predict the future concentration of air pollutants (APS) in Macau. Additionally, meteorological data and data on the concentration of APS have been utilized. Moreover, in Macau, some air quality monitoring stations (AQMSs) have less observed data in quantity, and, at the same time, some AQMSs recorded less observed data of certain types of APS. Therefore, the transfer learning and pre-trained neural networks have been employed to assist AQMSs with less observed data to build a neural network with high prediction accuracy. The experimental sample covers a period longer than 12-year and includes daily measurements from several APS as well as other more classical meteorological values. Records from five stations, four out of them are AQMSs and the remaining one is an automatic weather station, have been prepared from the aforesaid period and eventually underwent to computational intelligence techniques to build and extract a prediction knowledge-based system. As shown by experimentation, LSTM RNNs initialized with transfer learning methods have higher prediction accuracy; it incurred shorter training time than randomly initialized recurrent neural networks.

[*] Corresponding author. Tel: +34 954556237; fax: +34 954557139. Postal address: Reina Mercedes AV. 41012, Seville, Spain. Email address: atallon@us.es (A.J. Tallón-Ballesteros)

# 1. Introduction

With the development of societies and industries, many countries and cities in the world have to face the problem of air pollution (AP), which has been bringing many undesirable effects on human health. Therefore, predicting the AP level in the cities and then publishing the severity of air pollution to the public is important. Air pollutants (APS) mainly come from burning fossil fuels. They are mainly encompassing sulphur dioxide ($SO_2$), nitrogen monoxide (NO), nitrogen dioxide ($NO_2$), carbon monoxide (CO), inhalable particles with diameters which are generally 10 micrometers and smaller ($PM_{10}$), fine inhalable particles with diameters which are generally 2.5 micrometers and smaller ($PM_{2.5}$), etc. Indeed, PM stands for airborne particulate matter and its study is now on the rise especially to the current problem of the climate change associated to the vehicle emission and free-fuel transport. As we all know, AP adversely affects people's health, especially children and the elderly; it will also make patients with respiratory diseases, such as asthma and bronchitis, or cardiovascular disease, worse. In addition, prolonged exposure to traffic-related air pollution may shorten life expectancy. Moreover, people who go through long-term exposure to vehicle-related AP may have their life expectancy shortened [1]. Xi Chen *et al.* [2] studied the relationship between $NO_2$, $SO_2$, and $PM_{10}$ concentrations and lung cancer mortality in several northern cities in China, as well as the relationship between these APS and patients with lung cancer. The statistical data they have researched show that the concentration of air pollutants in people's area is positively correlated with the prevalence and mortality of lung cancer.

Atmosphere state and AP have a great relationship; for example, when the atmosphere is stable, that is to say, when the air in a certain area is not rising, the APS will stay on the surface, which is unfavorable to the spread of air pollutants. On the contrary, if the atmosphere is unstable, the air will move upward vertically, which will help the APS to spread to the sky. The atmosphere state is usually measured with seven different elements, namely wind speed, wind run, atmospheric temperature, relative humidity, dew point temperature, atmospheric pressure and precipitation. People usually employ automatic weather stations (AWSs) – also called meteorological monitoring stations – to measure automatically and periodically the above-mentioned seven atmospheric elements. Besides, air quality monitoring stations (AQMSs) are used to measure the concentration of APS such as $PM_{2.5}$, $SO_2$, NO, etc. in a certain area automatically and periodically. Magnitudes measured by AWSs and AQMSs along with their units are listed in Table 1.

Figure 1 (a) displays the locations of AQMSs in Macao [3], the following AQMSs have been used in this paper: High density residential area (Macao), Roadside (Macao), Ambient (Taipa), High density residential area (Taipa). Figure 1 (b) shows the locations of AWSs in Macao. Taipa Grande AWS has been utilized in this research. The observed data of the AQMSs and AWS in the circles with bullets in green and blue depicted in Figure 1 are used for the experimentation of this paper.

AP seriously affects public health. By letting people know the AP level such as $PM_{2.5}$, $SO_2$ and $NO_2$ in advance, they gain advantage and then they can plan for outdoor activities conveniently and protect people's health. Long-short term memory recurrent neural networks (LSTM RNNs) are used in this paper to predict the situations of AP in the future. LSTM RNNs are good at predicting time series data, and the concentration of APS can be considered as time series data. Hence, in order to be able to predict the already mentioned concentration of APS – despite the lack of knowledge about the atmospheric dispersion modeling of APS – LSTM RNN has been applied in this paper.

Additionally, in order to obtain good prediction results even though the lack of observed data, transfer learning has been proposed to be used in this paper to assist predicting the AP level. The LSTM RNNs have been trained in a domain (source domain) with more observed data as usual in data mining and knowledge-based systems, and then use the trained networks for the tasks with less observed data (target domain).

The objective of this paper is to investigate the difference, in terms of prediction errors, between our proposed method and the original method. The original method is to randomly initialize a neural network to do the prediction. The proposed method is to pre-train a neural work using transfer learning, with similar data from nearby stations that have certain correlation with the predicted results. The research question is whether or not to use pre-trained neural network methods.

The remaining of this paper is arranged as follows. Related works are introduced in Section 2. In Section 3, the methodology along with the key ingredients such as the LSTM RNNs, transfer learning and pre-trained neural networks are described. In Section 4, the details of the experiments are explained. Section 5 reports the experimental results and makes an analysis of experimental results. Section 6 draws the main conclusion as a summary of the experiments in the paper, opening some future lines for further works that can be extended.

| name of features | units |
|---|---|
| **air quality monitoring station** | |
| inhalable particles (PM10) | $\mu g/m^3$ |
| fine inhalable particles (PM2.5) | $\mu g/m^3$ |
| nitrogen monoxide (NO) | ppb |
| Nitrogen dioxide (NO2) | ppb |
| carbon monoxide (CO) | ppm |
| **automatic weather station** | |
| wind speed | km/h |
| wind run | km |
| station atmospheric pressure | hPa |
| air temperature | °C |
| relative humidity | % |

| precipitation | mm |
| dew point | °C |

**Table 1**. Features observed by Air Quality Monitoring Station (AQMS) and Automatic Weather Station (AWS).

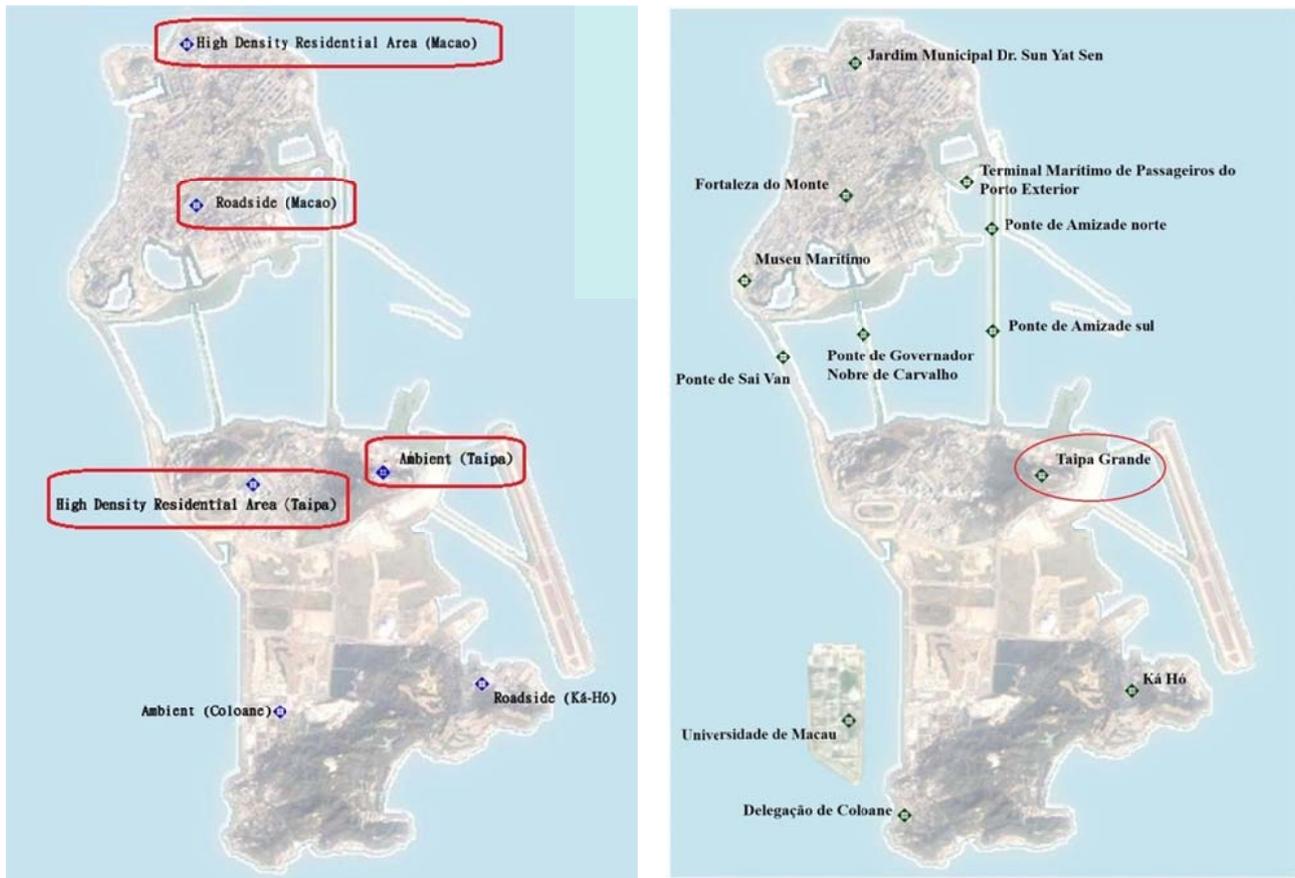

**Figure 1**. (a) Air quality monitoring stations (AQMSs) and (b) automatic weather stations (AWSs) in Macau.

## 2. Background

Transfer may have many meanings as well as application domains. On the one hand, from an engineering approach, a heat transfer from an object to another one may occur; additionally in the communication field, the transfer could be done in different ways like asynchronous, synchronous or even half-duplex or full-duplex mode. On the other hand, in artificial intelligence scope the learning may be transferred from a task to a subsequent task; in this way, the output of the first step is the input for the second step. Although transfer learning was introduced in the mid of the nineties from the previous century, it has not received a growing interest till 2018. The first special issue published on an international journal is dated in 1996 where a very extensive survey of transfer between connectionist networks was included [4]. Moreover, in the next year, that is 1997, a special issue attracted the attention from many researchers to dive into the inductive transfer [5]. By its part, Sebastian Thrun and Lorien Pratt edited a book entitled "Learning to Learn" in 1998 and Transfer Learning is addressed from four different points of views such as Overview, Prediction, Relatedness and Control including thirteen chapters [6]; humans often generalise correctly after a few of training examples by transferring knowledge acquired in other tasks; systems that learn to learn mimic this ability. During the period from 1998 to 2006, the topic remained in background. From 2007, it has been recognised as an important theme within machine learning. Basically, it is what happens when someone finds it much easier to learn to play chess having already learned to play checkers, or to recognize tables having already learned to recognize chairs [7].

Formally, transfer learning aims at providing a framework to make use of previously-acquired knowledge to solve new but similar problems much more quickly and effectively; in contrast to classical machine learning methods, transfer learning approaches exploit the knowledge accumulated from data in ancillary domains to facilitate predictive modelling consisting of different data patterns in the current domain [8]. Continuing with the transfer learning timeline, Sinno Jian Pan and Qiang Yang published a survey about the topic in 2009, and they studied extensively the inductive, transductive and unsupervised transfer learning [9]. The following year, the Handbook of Research in Machine Learning Applications devoted a chapter written by Lisa Torrey and Jude Shavlik to Transfer Learning to cover the inductive typology focusing on inductive, Bayesian and Hierarchical Transfer as well as the missing data and class labels; perhaps the main novelty of this work are the relationship with the reinforcement learning and the automatically mapping tasks [10]. A publication in the scope of forecasting, more concretely in the field of crude oil price, saw the light in 2012 [11]. The current journal published the first manuscript on transfer learning in 2015 introducing a transfer component analysis [12]; a newly paper falling in the survey category appeared also in 2015 which made emphasis on the computational intelligence corner to do the transfer [8]. To follow, Karl Weiss *et al.* wrote a survey paper with more than 140 references and a length of 40 pages in 2016 where some formal definitions are provided and a very wide taxonomy of many types of transfer learning, such as the homogeneous and the heterogeneous ones including, for each one, the asymmetric feature-based and symmetric feature-based transfer learning; in the former the parameter-based one, the relational-based one and the hybrid-based one are explained [13]. Stephan Spiegel opened a research line for Time Series Classification in dissimilarity spaces in 2016 [14]. Furthermore, the temporal information was also considered by Joseph Lemley *et al.* in the following year in the context of driver action classification [15]. Some months later, Ran Zhang *et al.* proposed neural networks to transfer the learning for bearing faults diagnosis in changing working conditions [16]. A good proliferation of works happened in 2018 and hence one may find insights for reconstruction and regression loss for time series [17] and frameworks based on extreme learning machine to conduct the transfer [18].

Jiangshe Zhang *et al.* [19] proposed an extreme learning machine [20] to predict the concentration of APS in two of locations in Hong Kong; they used a six-year air pollutant concentration and meteorological data from Hong Kong. They predicted APS comprising $NO_2$, $NO_X$, $O_3$, $SO_2$ and $PM_{2.5}$. Ming Cai *et al.* [21] utilized artificial neural network to forecast the hourly average concentration of APS in an arterial road in Guangzhou; the authors used concentration data of APS, meteorological and traffic video data, etc. to predict future concentrations of APS covering $CO$, $NO_2$, $PM_{10}$ and $O_3$. G. Grivas *et al.* [22] employed artificial neural network to predict the $PM_{10}$ concentration in Athens, Greece. S.I.V. Sousa *et al.* [23] applied principal component analysis [24] to pre-process the input data and then put it into an artificial neural network to predict the $O_3$ concentration Xiao Feng *et al.* [25] made use of a variety of techniques, including air mass trajectory analysis [26] and wavelet transformation [27] to pre-process times series data, and then incorporated the training data into artificial neural network to estimate the $PM_{2.5}$ concentration. Because *of* the large variability of $PM_{2.5}$ concentration in this area, the researchers used wavelet transformation to decompose the time series for $PM_{2.5}$ and then obtained several time series with less variability; then each time series is inputted to the neural network, and finally the prediction results of each neural network were combined together to obtain the $PM_{2.5}$ prediction result. Yu Zheng *et al.* [28] proposed a semi-supervised learning approach based on a co-training framework which consists of two separated classifiers, one is a spatial classifier based on an artificial neural network, and the other one is a temporal classifier based on a linear-chain Conditional Random Field (CRF). Bun Theang Ong *et al.* [29] applied deep recurrent neural network to predict $PM_{2.5}$ in Japan and employed auto-encoder as a pre-trained method to improve performance of deep recurrent neural network. Asha B Chelani *et al.* [30] utilized artificial neural network to measure the $SO_2$ concentration in three cities in Delhi, they used the Levenberg–Marquardt algorithm to train artificial neural networks.

**3. Proposed methodology**

In this work, the prediction of the concentration of APS is considered as a time series prediction problem and the LSTM RNNs is good for the time series prediction. Therefore, observed data from AWSs and AQMSs in Macao, and LSTM RNNs are used to predict the level of air pollution in the future. All AQMSs in Macau have officially begun to measure the concentration of $PM_{2.5}$ in July 2012. Therefore, the amount of $PM_{2.5}$ observed data for each station is relatively smaller than other APS. For some reasons, some AQMSs will have more observed data, whilst some stations will have fewer observed data. For example, the High density residential area (Taipa) AQMS suspended the air quality measurement from July 2012 to June 2013 due to the constructional engineering of the station nearby. Hence, it is suitable to transfer the knowledge of the RNNs of AQMSs with more observation data to the RNNs of AQMSs with fewer observation data. In our proposed design, the LSTM RNNs are used together to predict the concentration of APS at an AQMS and transfer learning methods are applied to train neural networks. At first, the LSTM RNNs are constructed and randomly initialized the weights of the LSTM; then observations of AWSs and AQMSs are used as training data for the neural network; and then the prediction ability of the neural network is evaluated. Then, the above-mentioned trained LSTM could be a pre-trained neural network for other predictive tasks, and transfer the knowledge to new neural networks in target domains. The new tasks may be predicting future concentration of certain APS – that can be the same or different from those of the source domain – in other AQMSs, that can be the same or different from source domain. In the new task, another LSTM is also constructed, and the weights of pre-trained neural network are used as the initial status of the new task. Then we put new task-related data to train the new LSTM. The above-mentioned transfer learning process is shown in the Figure 2.

The training data includes the observed data from various AQMSs and an AWS in Macau. The predicted target is the concentration of a certain air pollutant in an AQMS which is in the training data. In this paper, the scenarios for using transfer learning are as follows: (1) Construction and training of a RNN in the source domain for a type of air pollutant (e.g. $PM_{10}$) at an AQMS (e.g. Ambient (Taipa)); then using this neural network as the pre-trained neural network for another air pollutant (e.g. $PM_{2.5}$) of the same AQMS. That is, the task of the target domain is prediction of $PM_{2.5}$ of the same AQMS. (2) Creation and training in the source domain for an air pollutant concentration (e.g. $PM_{10}$) of an AQMS (e.g. Ambient (Taipa)). Then on the task of the target domain, the above mentioned RNN becomes a pre-trained network for another air pollutant (e.g. $PM_{2.5}$) of another AQMS (e.g. Roadside (Macau)); it is important to remark that the distributions are very similar in both aforementioned tasks. The different situations of the above transfer methods, aim at transferring the knowledge of neural networks with more training data to the tasks with fewer training data. In our prediction, LSTM RNNs are trained to predict the concentration of APS. According to the article of Lisa Torrey et al. [31], transfer learning can lead to better initial training status, faster learning speed and higher prediction accuracy. Therefore, it was expected that transfer learning would bring the above-mentioned benefits for the training process and result of LSTM RNNs. On the other hand, some AQMSs have less observed data, so the transfer learning is applied in this paper and eventually RNNs obtained good training results even in the case of less training data.

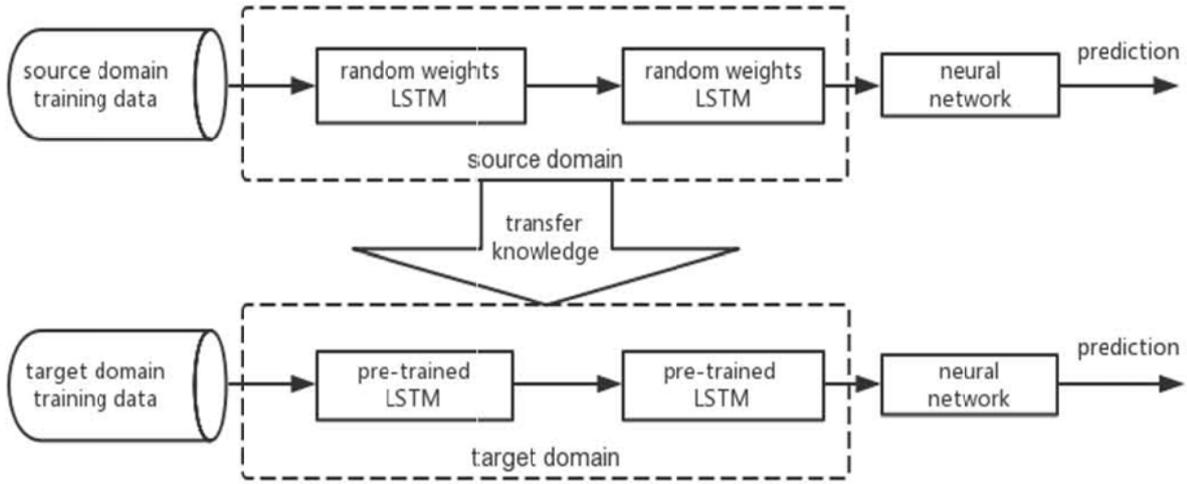

**Figure 2.** The process of transfer learning.

**4. Experimentation setup**

The problem to be solved in this paper is described as follows. There is a time series dataset $D$, $D = \{d_1, d_2, …, d_N\}$, where $i=1, …, N$, $N$ is the total number of records that the dataset has, and $d_i = (t_i, x_i)$, where $t_i$ is timestamp, it represents a certain time interval. The interval can be 24 hours, 1 hour, 1 minute, etc.; in this paper, the time interval is 1 day. In addition, $x_i$ contains meteorological data of an AWS, as well as data of concentration of APS from multiple AQMSs. Moreover, the predicted concentration of APS is also included in $x_i$ and these data are obtained at $t_i$ timestamp. In this experiment, when there is a certain time $t$, and there are $B$ time steps of observed data before $t$, the observed data of $B$ days before are used to predict the concentration value of air pollutant in timestamp $t$, as the Figure 3 shows.

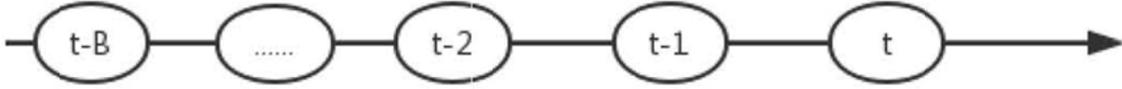

**Figure 3**. Observed data from the past several days was used to predict future air pollutants.

The trained RNN is represented by *p* (.), and the concentration value of air pollutant that is predicted by *p* (.) is represented by $y_{h,i}$, as shown in formula (1).

$$y_{h,i} = p(d_i) \tag{1}$$

In addition, the time series data consists of the observation records from the AWS and AQMSs in Macau, and once the time series data has been properly converted, it can become training, validation and test data for RNNs. The above data was obtained by submitting the application form to SMG [32]. What we want to predict is the concentration of APS of several AQMSs in time series data. The following is an example of conversion from time series data to LSTM training data. As shown in Figure 4, suppose there is a time series dataset *D* of length *N* and set *B*=6, then the time series dataset is converted to the following format: ($x_1$, $x_2$, $x_3$, $x_4$, $x_5$, $x_6$, $y_7$), ($x_2$, $x_3$, $x_4$, $x_5$, $x_6$, $x_7$, $y_8$), …, ($x_{n-6}$, $x_{n-5}$, $x_{n-4}$, $x_{n-3}$, $x_{n-2}$, $x_{n-1}$, $y_n$); in this abstract example, each training data is set to use the data 6 days before, and predict the air pollutant concentrations on the next day. Moreover, the above examples are the data conversion settings for each experiment in this paper.

| timestamp | x | y | timestamp | x | y | timestamp | x | y | timestamp | x | y | |
|---|---|---|---|---|---|---|---|---|---|---|---|---|
| 2003-01-01 | $x_1$ | $y_1$ | 2003-01-01 | $x_1$ | $y_1$ | 2003-01-01 | $x_1$ | $y_1$ | 2003-01-01 | $x_1$ | $y_1$ | |
| 2003-01-02 | $x_2$ | $y_2$ | 2003-01-02 | $x_2$ | $y_2$ | 2003-01-02 | $x_2$ | $y_2$ | 2003-01-02 | $x_2$ | $y_2$ | |
| 2003-01-03 | $x_3$ | $y_3$ | 2003-01-03 | $x_3$ | $y_3$ | 2003-01-03 | $x_3$ | $y_3$ | 2003-01-03 | $x_3$ | $y_3$ | |
| 2003-01-04 | $x_4$ | $y_4$ | 2003-01-04 | $x_4$ | $y_4$ | 2003-01-04 | $x_4$ | $y_4$ | 2003-01-04 | $x_4$ | $y_4$ | |
| 2003-01-05 | $x_5$ | $y_5$ | 2003-01-05 | $x_5$ | $y_5$ | 2003-01-05 | $x_5$ | $y_5$ | 2003-01-05 | $x_5$ | $y_5$ | |
| 2003-01-06 | $x_6$ | $y_6$ | 2003-01-06 | $x_6$ | $y_6$ | 2003-01-06 | $x_6$ | $y_6$ | 2003-01-06 | $x_6$ | $y_6$ | |
| 2003-01-07 | $x_7$ | $y_7$ | 2003-01-07 | $x_7$ | $y_7$ | 2003-01-07 | $x_7$ | $y_7$ | 2003-01-07 | $x_7$ | $y_7$ | |
| 2003-01-08 | $x_8$ | $y_8$ | 2003-01-08 | $x_8$ | $y_8$ | 2003-01-08 | $x_8$ | $y_8$ | 2003-01-08 | $x_8$ | $y_8$ | |
| 2003-01-09 | $x_9$ | $y_9$ | 2003-01-09 | $x_9$ | $y_9$ | 2003-01-09 | $x_9$ | $y_9$ | 2003-01-09 | $x_9$ | $y_9$ | |
| 2003-01-10 | $x_{10}$ | $y_{10}$ | 2003-01-10 | $x_{10}$ | $y_{10}$ | 2003-01-10 | $x_{10}$ | $y_{10}$ | 2003-01-10 | $x_{10}$ | $y_{10}$ | |
| … | … | … | … | … | … | … | … | … | … | … | … | … |
| $t_{n-2}$ | $x_{n-2}$ | $y_{n-2}$ | $t_{n-2}$ | $x_{n-2}$ | $y_{n-2}$ | $t_{n-2}$ | $x_{n-2}$ | $y_{n-2}$ | $t_{n-2}$ | $x_{n-2}$ | $y_{n-2}$ | |
| $t_{n-1}$ | $x_{n-1}$ | $y_{n-1}$ | $t_{n-1}$ | $x_{n-1}$ | $y_{n-1}$ | $t_{n-1}$ | $x_{n-1}$ | $y_{n-1}$ | $t_{n-1}$ | $x_{n-1}$ | $y_{n-1}$ | |
| $t_n$ | $x_n$ | $y_n$ | $t_n$ | $x_n$ | $y_n$ | $t_n$ | $x_n$ | $y_n$ | $t_n$ | $x_n$ | $y_n$ | |

**Figure 4**. Time series data is converted to training data for RNN.

The architecture of all randomly initialized neural networks is shown in Figure 5. It is a concrete example which we explain now; once all the features from AQMSs and AWS are combined, in a timestamp, there are 41 columns given by AWS and AQMSs, and each row of the dataset used data 6 days ago from each station.

Figure 6 is an example about architecture of LSTM RNN in target domain. The top of RNNs in the source domain are added the RNN layers that match the feature space of the target domain; the total number of layers of neural networks in target domain is constant, but the number of neurons on the input layer and 2$^{nd}$ layers will be changed according to the feature space that in target domain.

In our work, some LSTM RNNs are randomly initialized their weights and biases matrices, then are trained and used to predict APS. Each air pollutant will have a corresponding RNN to predict its concentration, the $PM_{10}$, $NO_2$, NO and $PM_{2.5}$ of the Ambient (Taipa) AQMS and the CO of the High density residential area (Macau) AQMS are predicted. The $PM_{2.5}$ concentration of all AQMSs in our case has observed data. So we will provide here pre-trained neural networks for RNNs which predict the concentration of $PM_{2.5}$ of all AQMSs, and the pre-trained neural networks of source domain include RNNs for predicting $PM_{10}$ of Roadside (Macau), $PM_{10}$ density residential area (Macau) and $PM_{10}$ of Ambient (Taipa), respectively. About High density residential area (Taipa) AQMS, this is a station with fewer observed data, compared to other AQMSs. For this AQMS, the predicted APS are $PM_{2.5}$, $NO_2$, NO and CO. Moreover, randomly initialized neural networks and pre-trained neural networks method are used for comparison to assess the performances and results of the two training methods. Another reason for using transfer learning is to use multiple pre-trained neural networks for predicting different APS in the same AQMS as the source domain, and then the target domain is predicting the $PM_{2.5}$ of the same AQMS to see how the transfer learning works.

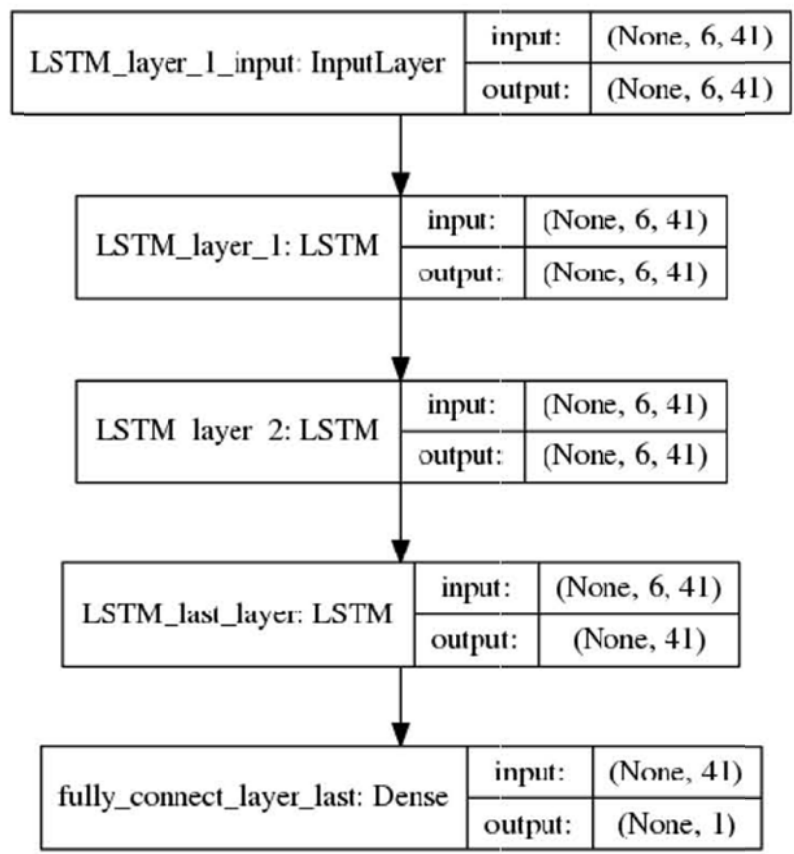

**Figure 5**. The architecture that is used by all randomly initialized neural networks.

The values of each feature in the dataset were rescaled, and then the observed data is put into the neural networks for training, validation and test. The formula (2) and (3) are the rescaling formulas, where $X_{min}$ is the minimum value in a certain feature, $X_{max}$ is the maximum value in a certain feature, $R_{min}$ is the minimum value after rescaled, and $R_{max}$ is the maximum value after rescaled. In this article, the rescaled ranges of all features are from 0 to 1.

$$X_{std} = \frac{X - X_{min}}{X_{max} - X_{min}} \quad (2)$$

$$X_{scaled} = X_{std} * (R_{max} - R_{min}) + R_{min} \quad (3)$$

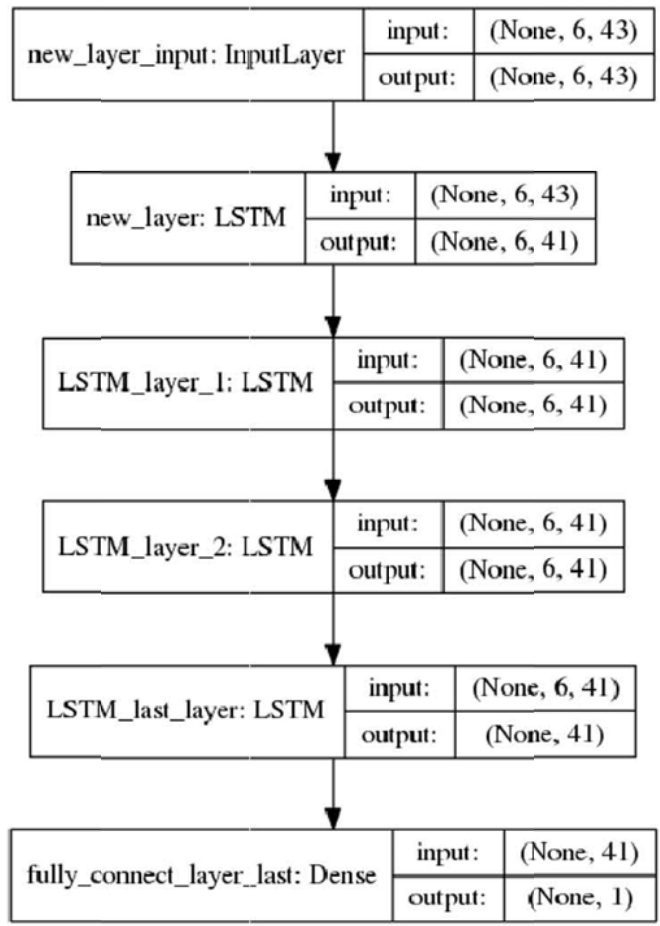

**Figure 6**. An example of RNN's architecture trained network methods.

In each experiment, the observed data was divided into three parts: the 1st part, approximately the 70% as training data; the 2nd part, 25% as the validation data; the 3rd part, 5% as the testing data. The study period encompasses more than 12 years of 2001-2014; the starting and finishing dates are 1st October 2001 and 1st July 2014, respectively, which cover 4,656 days with a daily sample and represent exactly 12.75 years. Hence, 9 years (71%) of samples are the training set, 3 years (25%) of samples act as validation set and 0.75 years (5%) are set as the testing set.

# 5. Experimental results and analysis

This section aims at presenting the results raised by the experimentation. Firstly, there is a comparison between the new approach and the original procedure. Secondly, an alternative procedure and the original approach are reported.

## 5.1. Base scenario

The base scenario pursues to investigate the difference, in terms of prediction errors, between our proposed method and the original method. The original method is to randomly initialize a neural network for doing the prediction. The proposed method is to pre-train a neural work using transfer learning, with similar data from nearby stations that have certain correlation with the predicted results. The experimental results, measured in MSE, using training and validation data for randomly initialized networks and pre-trained networks are tabulated in Table 2. The training results for various networks, including randomly initialized neural networks that predicted $PM_{2.5}$ for various AQMSs, and neural networks for various AQMSs that predict $PM_{2.5}$ based on the $PM_{10}$ pre-trained neural networks at all AQMSs are reported in Table 2. It can be seen that the neural networks that used pre-trained methods, and used the pre-trained network which is predicting the same (or similar for $PM_{2.5}$) air pollutants and at different AQMSs, can indeed provide higher accuracy, faster learning speed and better initial learning state for neural networks.

| source domain | target domain | training data | | | validation data | | |
|---|---|---|---|---|---|---|---|
| | | Best MSE | Best epoch | Initial MSE | Best MSE | Best epoch | Initial MSE |
| N/A | Roadside (Macau) PM2.5 | 0.007266084 | 400 | 0.061714470 | 0.004698436 | 242 | 0.045378533 |
| Roadside (Macau) PM10 | | 0.005730863 | **242** | 0.025669344 | 0.003596468 | **92** | 0.049579145 |
| High density residential area (Macau) PM10 | | **0.005546452** | 261 | **0.024579911** | 0.003505309 | 117 | **0.043278625** |
| Ambient (Taipa) PM10 | | 0.005549194 | 245 | 0.025933107 | **0.003466034** | 95 | 0.043425080 |
| N/A | High density residential area (Macau) PM2.5 | 0.006227937 | 303 | 0.055660227 | **0.002996397** | 153 | 0.028110390 |
| Roadside (Macau) PM10 | | 0.004939886 | 168 | 0.021892348 | 0.003607673 | **18** | 0.031786795 |
| High density residential area (Macau) PM10 | | **0.004480842** | 188 | **0.021647959** | 0.003216892 | 38 | **0.026681633** |
| Ambient (Taipa) PM10 | | 0.004656768 | 171 | 0.021687523 | 0.003423001 | 21 | 0.027333746 |
| N/A | Ambient (Taipa) PM 2.5 | 0.007895702 | 368 | 0.059606799 | 0.007581763 | 197 | 0.045722324 |
| Roadside (Macau) PM10 | | 0.006281779 | 197 | 0.027147979 | 0.005756136 | 47 | 0.051130223 |
| High density residential area (Macau) PM10 | | 0.005962373 | **186** | 0.027097617 | **0.005709018** | **36** | **0.040284840** |
| Ambient (Taipa) PM10 | | **0.005713619** | 253 | **0.026958865** | 0.005752293 | 103 | 0.048785915 |

**Table 2**. Comparison of experimental results using training and validation data for randomly initialized networks and pre-trained networks ($PM_{2.5}$).

To illustrate the results, some charts are depicted in this second part of the paragraph. The research question is whether or not after using pre-trained neural network methods, certain benefits may be achieved as follows: (1) better initial state; (2) fewer epochs are required in training for convergence; and (3) better predictive ability is reached. Figures 7, 9, 11, 12 show the comparison of real (observed data) and predicted values. This set of charts shows that generally when the concentrations of APS were low, the predictions were accurate. The reason is that the high concentration of air pollutants are outliers and RNNs were not designed to deliberately cope with outliers during the training process. It can be seen from the other set of charts, depicted in Figures 8, 10, 13, which are trends of loss functions that use training data; in most cases, in terms of Best MSE and the number of epochs required to obtain the Best MSE, using pre-trained neural networks were better than random initialized neural networks. Figure 10 is tied to Figure 6 given that the architecture of the latter corresponds to the former scenario.

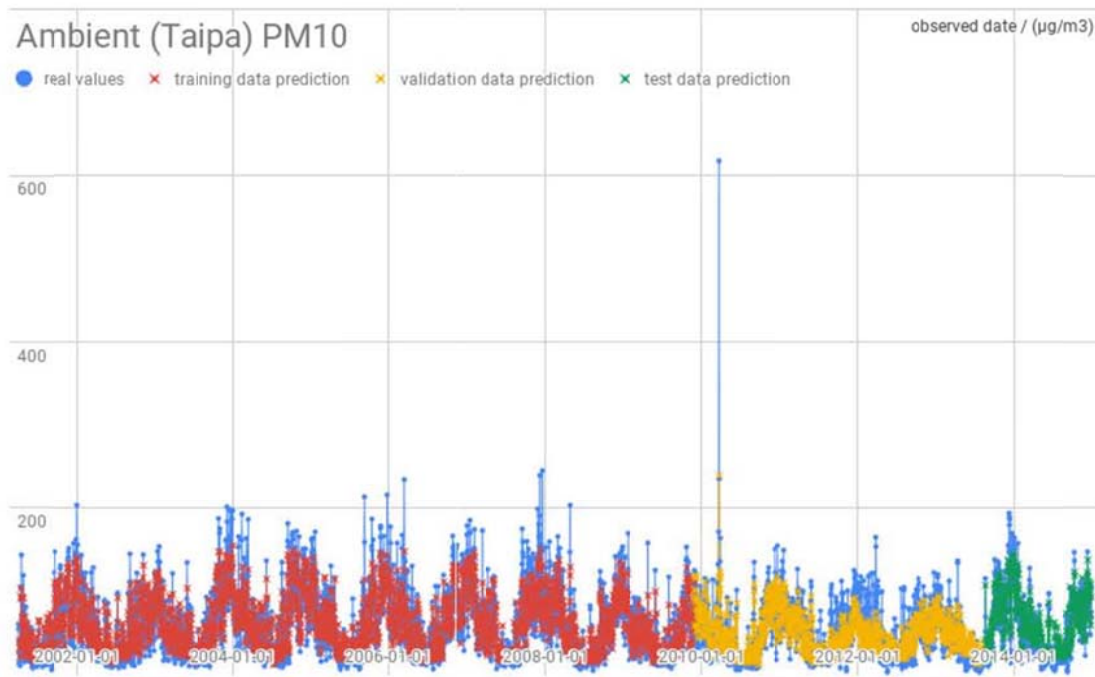

**Figure 7**. Comparison of real and predicted values in Ambient (Taipa) for $PM_{10}$.

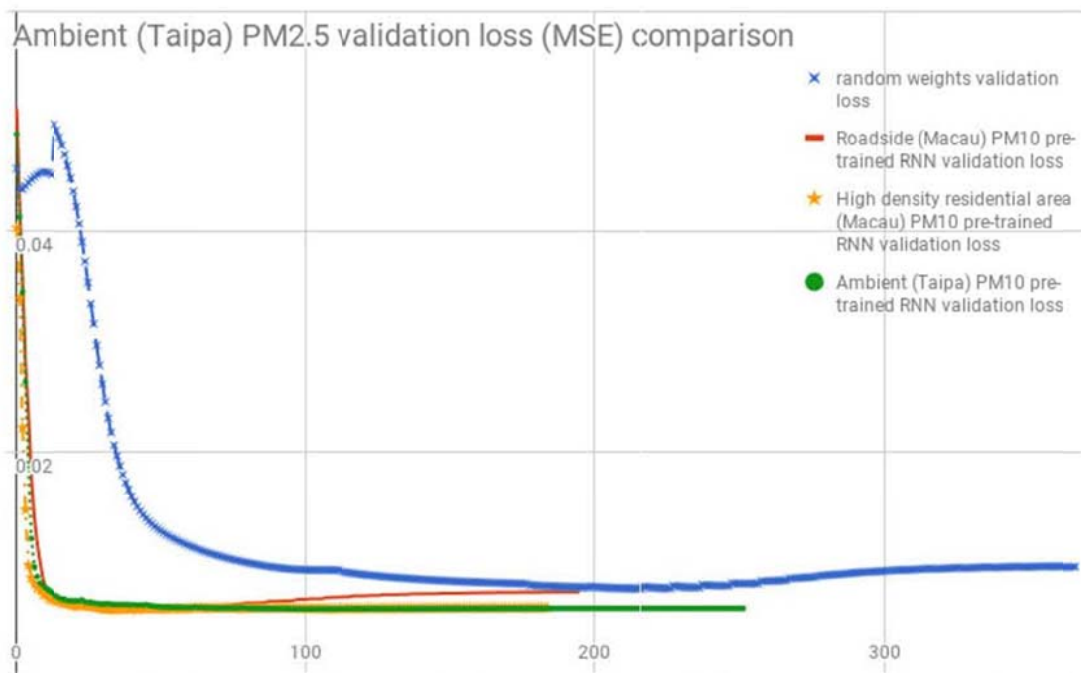

**Figure 8**. Comparison of the trends of loss functions (used validation data, Ambient (Taipa) $PM_{2.5}$) – pre-trained with data from multiple AQMSs $PM_{10}$ to Ambient (Taipa) $PM_{2.5}$

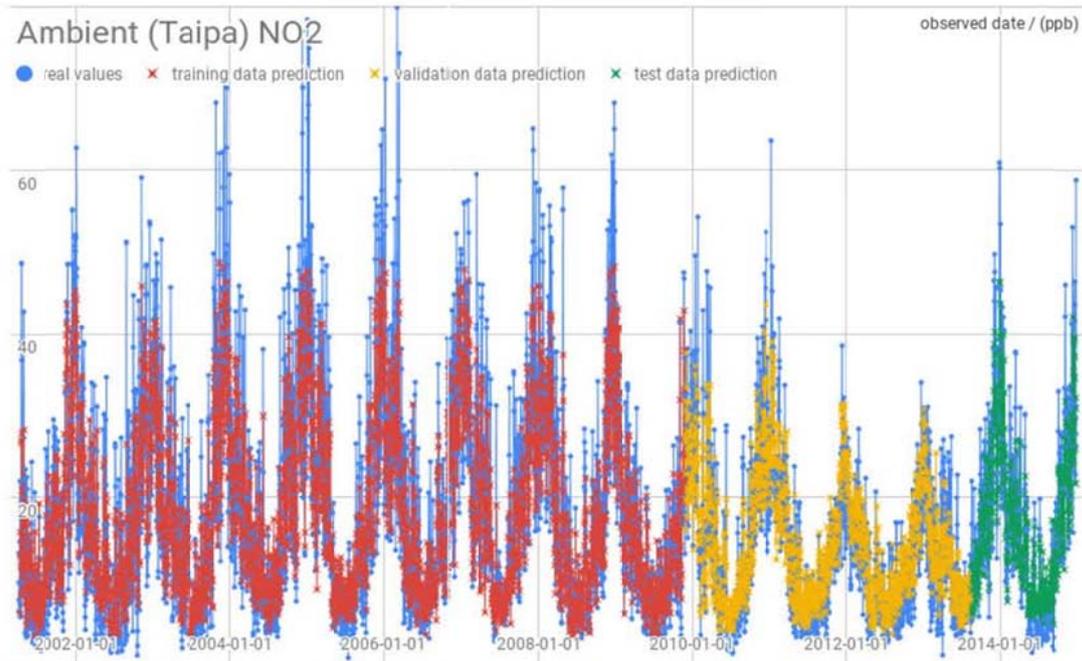

**Figure 9**. Comparison of real and predicted values in Ambient (Taipa) for $NO_2$.

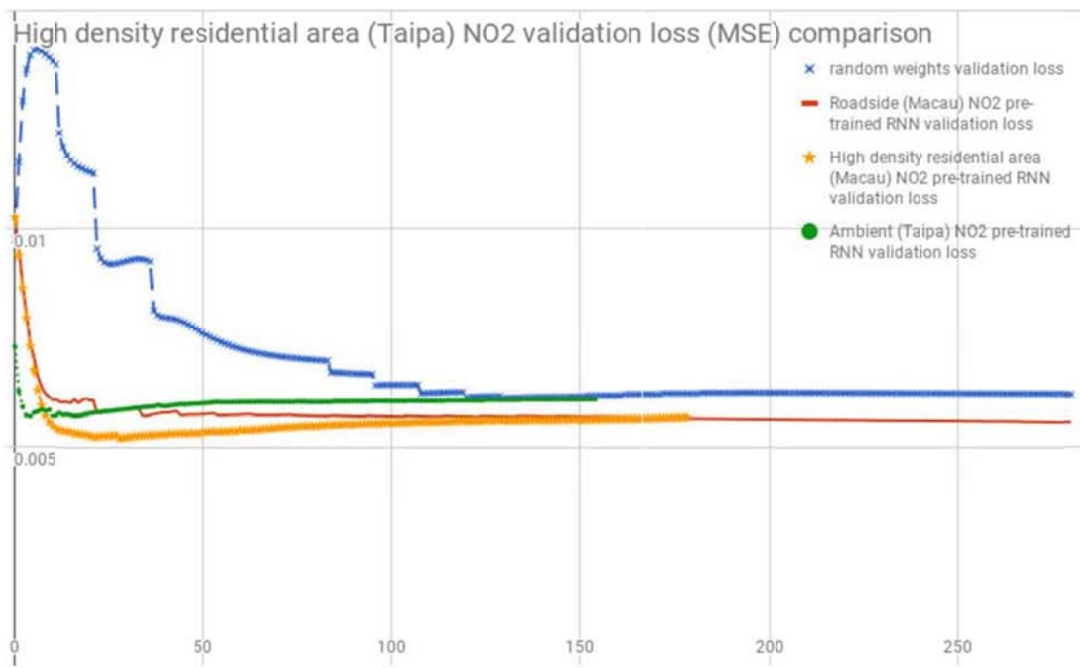

**Figure 10**. Comparison of the trends of loss functions of High density residential area (Taipa) $NO_2$ – pre-trained from AQMSs $NO_2$ to High density residential area (Taipa) $NO_2$.

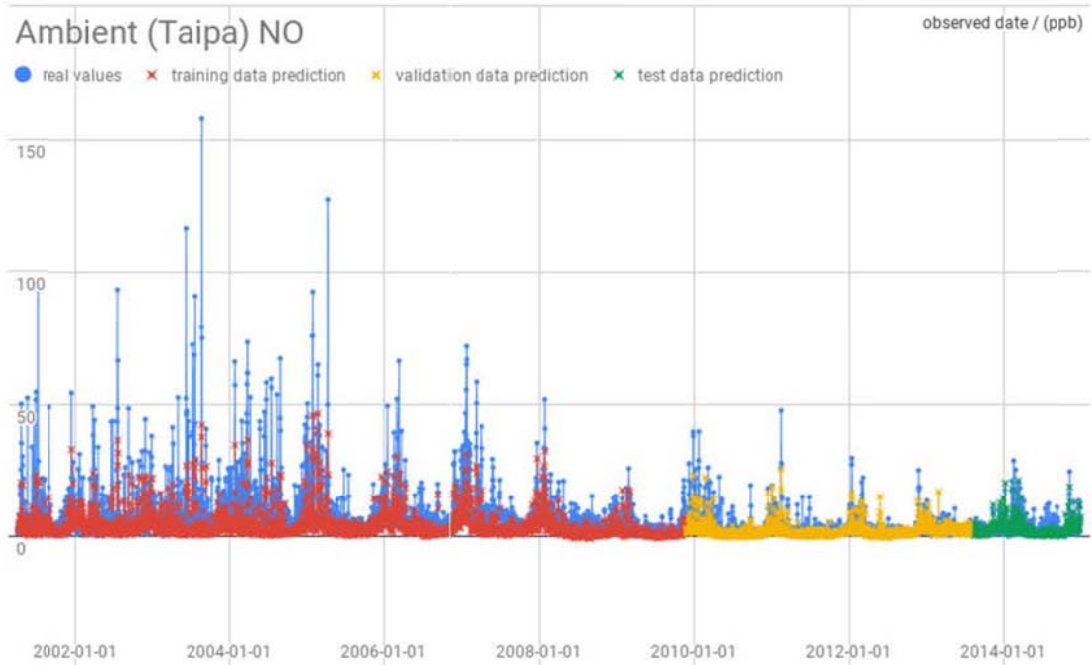

**Figure 11**. Comparison of real and predicted values in High density residential area (Taipa) for NO.

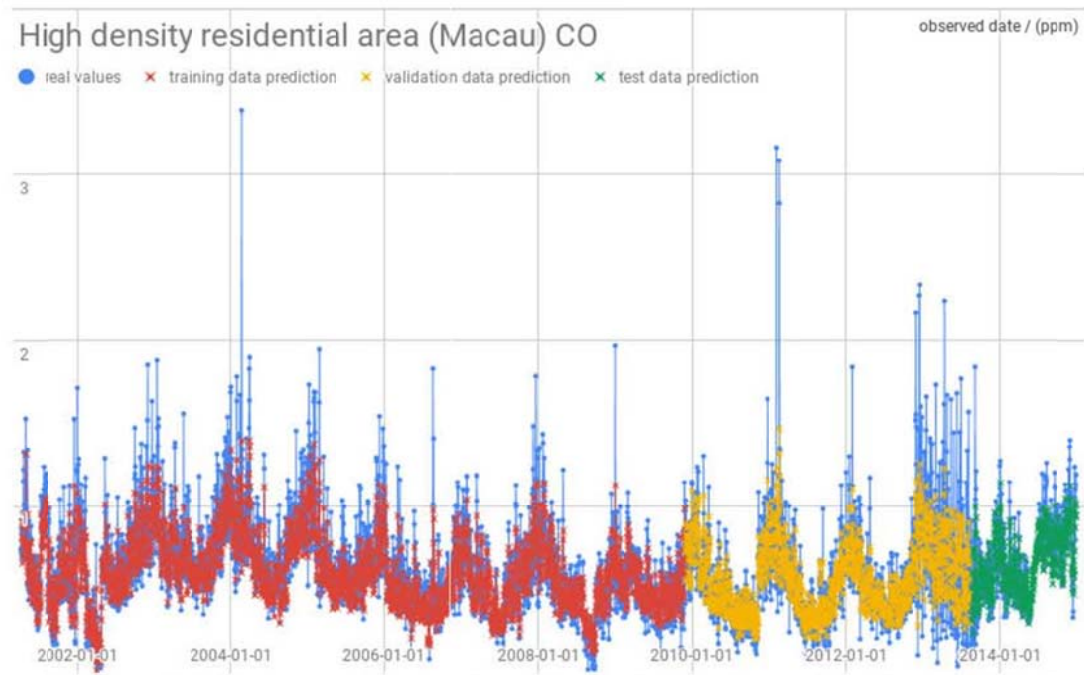

**Figure 12**. Comparison of real and predicted values in High density residential area (Taipa) for CO).

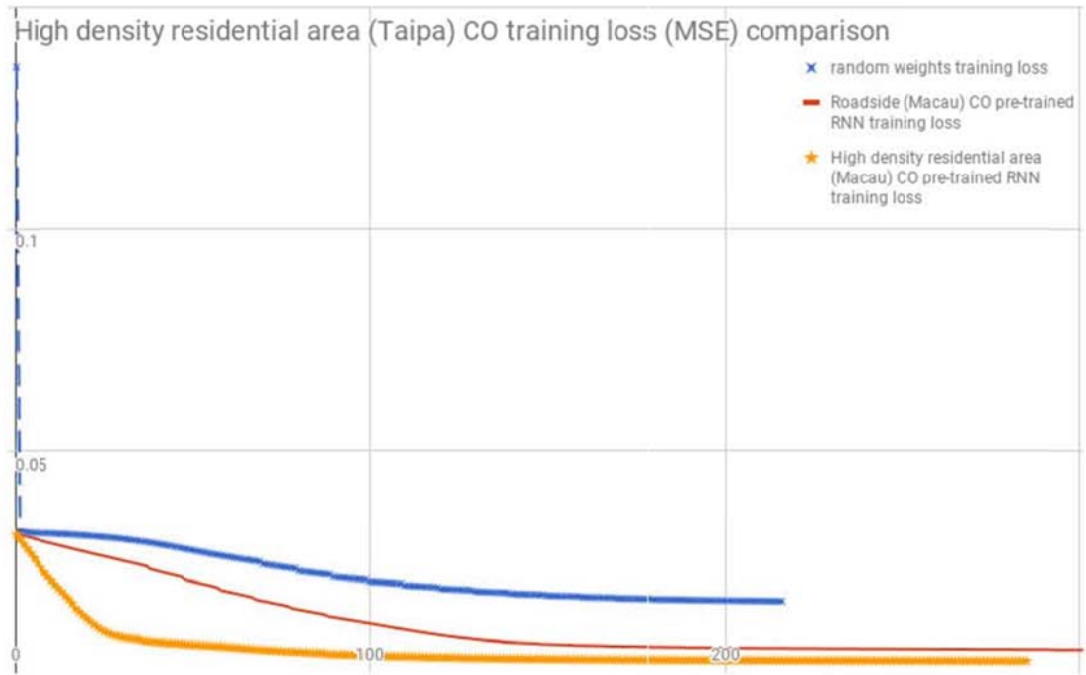

**Figure 13**. Comparison of the trends of loss functions of High density residential area (Taipa) CO – pre-train from AQMSs CO to High density residential area (Taipa) CO.

**5.2. Alternative scenario**

The alternative scenario proposes the use of the input layer of the target domain in two ways: trainable and untrainable. In the first case, the source domain layer is set to trainable and in the second case, the pre-trained LSTM can be trained in the target domain. Figure 14 depicts the alternative scenario.

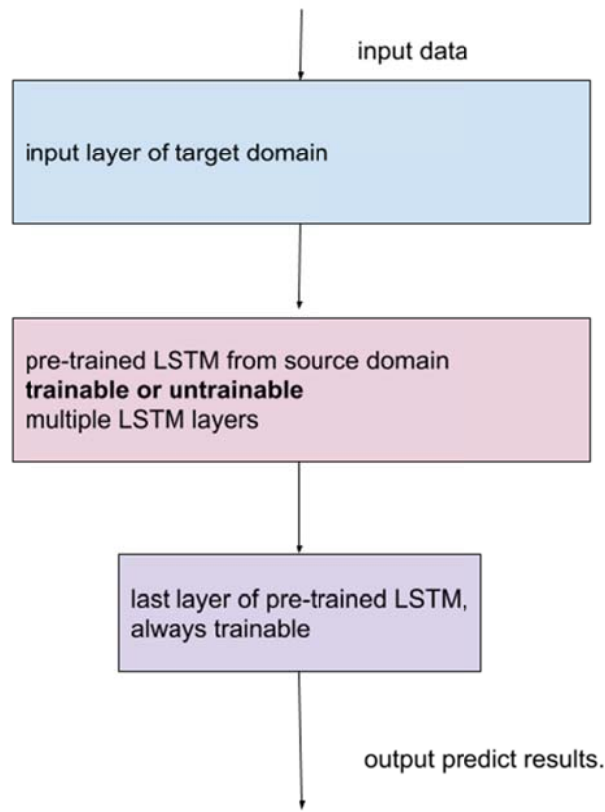

**Figure 14**. Chart of the alternative scenario including the target domain.

The results concerning the randomly initialized networks and the alternative scenario are shown in Table 3. The experimental results reveal that in most cases, the prediction results using pre-trained LSTMs, including trainable and untrainable pre-trained LSTMs, will be better than randomly initialized LSTMs. The best results for each target domain between every pair of trainable and untrainable, as well as the random initialization source domains are highlighted in bold font. The computational cost of random initialization is in the order the seconds and trainable and untrainable networks require a few minutes to be processed completely.

| source domain | target domain | training data | | | validation data | | |
|---|---|---|---|---|---|---|---|
| | | Best MSE | Best epoch | Initial MSE | Best MSE | Best epoch | Initial MSE |

| Pre-training | Training/Validation | | | | | | |
|---|---|---|---|---|---|---|---|
| N/A | | 0.007266084 | 400 | 0.061714470 | 0.004698436 | 242 | 0.045378533 |
| Roadside (Macau) PM10, pre-trained LSTM is trainable | | **0.005182960193** | 227 | **0.02527686319** | **0.00358629241** | 77 | 0.04863139938 |
| Roadside (Macau) PM10, pre-trained LSTM is untrainable | | 0.005598728544 | **208** | 0.02950307199 | 0.004178910752 | **60** | **0.02185823754** |
| High density residential area (Macau) PM10, pre-trained LSTM is trainable | Roadside (Macau) PM 2.5 | **0.004859869157** | 256 | **0.02420120651** | **0.003472965484** | 106 | 0.04286825121 |
| High density residential area (Macau) PM10, pre-trained LSTM is untrainable | | 0.005239599353 | **239** | 0.0265777999 | 0.003741108357 | **89** | **0.01998857462** |
| Ambient (Taipa) PM10, pre-trained LSTM is trainable | | **0.005092177982** | 225 | **0.02558391114** | **0.003491097135** | 75 | 0.04165692058 |
| Ambient (Taipa) PM10, pre-trained LSTM is untrainable | | 0.00530908497 | 260 | 0.02893763905 | 0.003937646276 | 110 | **0.02238395186** |
| N/A | | 0.006227937 | 303 | 0.055660227 | **0.002996397** | 153 | 0.028110390 |
| Roadside (Macau) PM10, pre-trained LSTM is trainable | | **0.004569249764** | 167 | **0.02211847892** | 0.003569208331 | 18 | 0.03125582054 |
| Roadside (Macau) PM10, pre-trained LSTM is untrainable | | 0.00499207167 | **160** | 0.02538478752 | 0.004096408385 | **10** | **0.01251311171** |
| High density residential area (Macau) PM10, pre-trained LSTM is trainable | High density residential area (Macau) PM2.5 | **0.004235766168** | 166 | **0.02143066662** | 0.003448928911 | 16 | 0.02638574313 |
| High density residential area (Macau) PM10, pre-trained LSTM is untrainable | | 0.004730619973 | **156** | 0.02372692178 | 0.003976781423 | **6** | **0.01128700101** |
| Ambient (Taipa) PM10, pre-trained LSTM is trainable | | **0.003870524781** | 282 | **0.03535970361** | 0.005476613939 | 132 | **0.01998347242** |
| Ambient (Taipa) PM10, pre-trained LSTM is untrainable | | 0.00464007655 | 313 | 0.04280769982 | 0.005482501299 | 164 | 0.02113966902 |
| N/A | | 0.007895702 | 368 | 0.059606799 | 0.007581763 | 197 | 0.045722324 |
| Roadside (Macau) PM10, pre-trained LSTM is trainable | | 0.005870847587 | **187** | **0.02695635663** | **0.005757841261** | 37 | 0.03692072487 |
| Roadside (Macau) PM10, pre-trained LSTM is untrainable | | **0.005667415343** | 204 | 0.03032668894 | 0.005965355129 | 54 | **0.02517349517** |
| High density residential area (Macau) PM10, pre-trained LSTM is trainable | Ambient (Taipa) PM 2.5 | 0.005870847587 | **187** | **0.02695635663** | **0.005757841261** | 37 | 0.03692072487 |
| High density residential area (Macau) PM10, pre-trained LSTM is untrainable | | **0.005667415343** | 204 | 0.03032668894 | 0.005965355129 | 54 | **0.02517349517** |
| Ambient (Taipa) PM10, pre-trained LSTM is trainable | | **0.005138671779** | 230 | **0.02699790372** | 0.005343422969 | 124 | **0.02564689896** |
| Ambient (Taipa) PM10, pre-trained LSTM is untrainable | | 0.005666261346 | 273 | 0.03044375491 | **0.005228251649** | 80 | 0.04519483775 |

**Table 3**. Comparison of experimental results for PM$_{2.5}$ using training and validation data for randomly initialized networks and pre-trained LSTM in two ways: trainable and untrainble.

## 6. Conclusions and future work

In this paper, we proposed a type of transfer learning model that combines LSTM RNNs for predicting air pollutant concentrations. The results from our experiments show that, pre-trained neural network methods are helpful for training neural networks. In other words, the LSTM RNNs that are initialized with pre-trained neural networks can achieve a higher level of prediction accuracy. Furthermore, the number of epochs that are required to train a LSTM RNNs into convergence can be reduced. The new method creates better initial states for RNNs. Our current experiments are concerned with predicting the values of air pollutant concentration on the next day. As future work, the proposed method can be used in training the RNNs to predict air pollution on the next several days ahead or even in a period shorter than one day. Moreover, hourly-observed data could be used to predict hourly data in the next several hours for enhancing the timely density of air pollution predictions.

The method proposed in the paper can also be used to predict other air pollutants, such as $O_3$ and $SO_2$. These air pollutants predictors can be modified so that the predictor can be trained with the latest observed data continuously. The diversity of the data could be expanded to more observed data of AWSs and AQMSs. Using other type of data, such as vehicle traffic data to predict concentration of APS of roadside by AQMSs, etc. should be attempted. The prediction border could be expanded too. Observed data of AWSs and AQMSs from Guangdong and Hong Kong can be used as training data instead of just using the data locally in Macau. Residents nowadays are more concerned about the situation of serious AP, that is, the situation of high concentration of APS. However, occurrence of serious AP is relatively rare and unusual, and even these conditions are considered abnormal values (outliers). Therefore, some imbalanced dataset processing methods can be used along with LSTM RNNs in future work, so that predictions can be made more accurate prior to the AP scenarios.

**Acknowledgement**

The authors are thankful to the financial support from the research grants, 1) MYRG2016-00069, titled 'Nature-Inspired Computing and Metaheuristics Algorithms for Optimizing Data Mining Performance' and 2) MYRG2016-00217, titled 'Improving the Protein-Ligand Scoring Function for Molecular Docking by Fuzzy Rule-based Machine Learning Approaches' offered by University of Macau and Macau SAR government.